\documentclass[11pt,a4paper]{article}
\usepackage[utf8]{inputenc}
\usepackage[T1]{fontenc}
\usepackage{amsmath,amssymb,amsthm}
\usepackage{graphicx}
\usepackage{booktabs}
\usepackage{hyperref}
\usepackage{algorithm}
\usepackage{algorithmic}
\usepackage[table]{xcolor}
\usepackage{enumitem}
\usepackage{multirow}
\usepackage[margin=1in]{geometry}
\usepackage{natbib}
\usepackage{subcaption}
\usepackage{tikz}
\usetikzlibrary{arrows.meta,positioning,calc,fit,backgrounds,decorations.pathreplacing,shapes.geometric}

\newtheorem{definition}{Definition}

\title{Outcome-Aware Tool Selection for Semantic Routers:\\Latency-Constrained Learning Without LLM Inference}

\author{%
  Huamin Chen$^{1}$ \quad
  Xunzhuo Liu$^{1}$ \quad
  Junchen Jiang$^{2}$ \quad
  Bowei He$^{3}$ \quad
  Xue Liu$^{3}$
  \\[6pt]
  $^{1}$vLLM Semantic Router Project \quad
  $^{2}$Tensormesh Inc / UChicago \quad
  $^{3}$MBZUAI / McGill University
}

\date{}

\begin{document}

\maketitle

\begin{abstract}
Semantic routers in LLM inference gateways select tools in the critical request path, where every millisecond of added latency compounds across millions of requests. We propose \textbf{Outcome-Aware Tool Selection (OATS)}, which interpolates tool embeddings toward the centroid of queries where they historically succeed---an offline process that adds no parameters, latency, or GPU cost at serving time. On MetaTool (199~tools, 4,287~queries), this improves NDCG@5 from 0.869 to 0.940; on ToolBench (2,413~APIs), from 0.834 to 0.848. We also evaluate two learned extensions: a 2,625-parameter MLP re-ranker and a 197K-parameter contrastive adapter. The MLP re-ranker hurts or matches baseline when outcome data is sparse relative to the tool set; the contrastive adapter provides comparable gains on MetaTool (NDCG@5: 0.931). All methods are evaluated on the same held-out 30\% test split. The practical takeaway is to start with the zero-cost refinement and add learned components only when data density warrants it. All mechanisms run within single-digit millisecond CPU budgets.
\end{abstract}

\section{Introduction}
\label{sec:intro}

LLM serving infrastructure increasingly uses \emph{semantic routers}---lightweight proxies that inspect requests and make routing decisions before forwarding them to backend model pools. These routers sit in the critical path: added latency compounds across millions of daily requests. Among their functions---model selection, prompt-length-based pool assignment, rate limiting---one of the most consequential for agentic applications is \emph{tool selection}: choosing which tools to attach to a request before it reaches the LLM.

\subsection{The Latency--Accuracy Tradeoff}

Tool selection in a router must satisfy a hard constraint:

\begin{quote}
\emph{Complete within single-digit milliseconds, without GPU resources or LLM inference.}
\end{quote}

\noindent At 10,000 requests/second with a 5\,ms budget, tool selection alone consumes 50 CPU-seconds per second. Any method requiring LLM inference---even on a small model---exceeds this budget by orders of magnitude. Table~\ref{tab:cost_comparison} shows that all OATS methods stay within this budget on CPU.

\begin{table}[h]
\centering
\caption{Cost of tool selection mechanisms. Latency is per-request p50 on 2,413 tools (ToolBench). All methods run on CPU.}
\label{tab:cost_comparison}
\begin{tabular}{l r r l l}
\toprule
\textbf{Method} & \textbf{Latency} & \textbf{Parameters} & \textbf{GPU} & \textbf{Viable at} \\
                 & \textbf{(ms)}    &                     & \textbf{Required} & \textbf{10K rps?} \\
\midrule
BM25 (lexical)     & $\sim$7       & 0     & No  & \checkmark \\
Static embedding   & $\sim$5       & 22M$^\dagger$ & No$^*$ & \checkmark \\
OATS-S1 (offline)  & $\sim$4       & 22M$^\dagger$ & No$^*$ & \checkmark \\
OATS-S2 (+ MLP)    & $\sim$5       & 22M + 2.6K    & No$^*$ & \checkmark \\
OATS-S3 (+ adapter)& $\sim$5       & 22M + 197K    & No$^*$ & \checkmark \\
\bottomrule
\multicolumn{5}{l}{\footnotesize $^\dagger$\,Embedding model (all-MiniLM-L6-v2); runs on CPU at batch throughput.} \\
\multicolumn{5}{l}{\footnotesize $^*$\,CPU inference; GPU optional for higher throughput but not required.}
\end{tabular}
\end{table}

Figure~\ref{fig:arch_comparison} illustrates the architecture difference. In the LLM-based approach, a dedicated orchestrator runs full inference on every request---requiring GPU resources and adding hundreds of milliseconds. In the semantic router approach, selection is a lightweight CPU operation, but it relies on static similarity with no learning. OATS preserves the fast path while incorporating outcome feedback through offline loops.

\begin{figure}[t]
\centering
\begin{tikzpicture}[
    >=Stealth,
    every node/.style={font=\small},
    box/.style={draw, rounded corners=3pt, minimum height=0.7cm, minimum width=1.4cm, align=center, thick},
    gpubox/.style={box, fill=red!12, draw=red!60!black},
    cpubox/.style={box, fill=blue!10, draw=blue!60!black},
    oatsbox/.style={box, fill=green!12, draw=green!50!black},
    timebar/.style={fill=#1, rounded corners=1pt, minimum height=0.25cm},
    lbl/.style={font=\scriptsize\sffamily, text=black!70},
    arw/.style={->, thick, color=black!70},
    scale=0.92, transform shape,
]

\node[font=\small\bfseries, anchor=west] at (-1.0, 1.0) {(a) LLM-based tool selection};

\node[cpubox] (user1) at (0, 0) {User};
\node[cpubox] (router1) at (2.2, 0) {Router\\[-1pt]{\tiny (proxy)}};
\node[gpubox, minimum width=2.6cm] (orch) at (5.6, 0) {LLM Orchestrator\\[-1pt]{\tiny 8B--70B params}};
\node[gpubox, minimum width=1.6cm] (llm1) at (9.0, 0) {LLM\\[-1pt]{\tiny Serving}};
\node[cpubox] (resp1) at (11.4, 0) {Response};

\draw[arw] (user1) -- (router1);
\draw[arw] (router1) -- (orch);
\draw[arw] (orch) -- (llm1) node[midway, above=5pt, lbl, fill=white, inner sep=1pt] {+ tools};
\draw[arw] (llm1) -- (resp1);

\draw[decorate, decoration={brace, amplitude=4pt, mirror}, red!60!black, thick]
    ([yshift=-0.5cm]orch.south west) -- ([yshift=-0.5cm]orch.south east)
    node[midway, below=5pt, font=\scriptsize\sffamily, text=red!70!black] {500--2000\,ms, GPU};

\draw[decorate, decoration={brace, amplitude=4pt, mirror}, red!60!black, thick]
    ([yshift=-0.5cm]llm1.south west) -- ([yshift=-0.5cm]llm1.south east)
    node[midway, below=5pt, font=\scriptsize\sffamily, text=red!70!black] {200--2000\,ms};

\node[timebar=red!25, minimum width=7.0cm, anchor=west] at ([yshift=-1.6cm]router1.south west) {};
\node[font=\scriptsize\sffamily, text=red!70!black, anchor=west] at ([yshift=-1.6cm, xshift=7.2cm]router1.south west) {\textbf{700--10,000\,ms}};

\node[font=\small\bfseries, anchor=west] at (-1.0, -3.5) {(b) OATS semantic router (ours)};

\node[cpubox] (user2) at (0, -4.5) {User};
\node[oatsbox, minimum width=4.2cm] (router2) at (3.8, -4.5) {%
    \begin{tabular}{c}
    Semantic Router {\tiny (CPU)} \\[-2pt]
    {\tiny embed $\to$ sim $\to$ re-rank $\to$ top-$K$}
    \end{tabular}};
\node[gpubox, minimum width=1.6cm] (llm2) at (8.2, -4.5) {LLM\\[-1pt]{\tiny Serving}};
\node[cpubox] (resp2) at (10.6, -4.5) {Response};

\draw[arw] (user2) -- (router2);
\draw[arw] (router2) -- (llm2) node[midway, above=5pt, lbl, fill=white, inner sep=1pt] {query + tools};
\draw[arw] (llm2) -- (resp2);

\draw[decorate, decoration={brace, amplitude=4pt, mirror}, green!50!black, thick]
    ([yshift=-0.5cm]router2.south west) -- ([yshift=-0.5cm]router2.south east)
    node[midway, below=5pt, font=\scriptsize\sffamily, text=green!50!black] {3--7\,ms, CPU only};

\draw[decorate, decoration={brace, amplitude=4pt, mirror}, red!60!black, thick]
    ([yshift=-0.5cm]llm2.south west) -- ([yshift=-0.5cm]llm2.south east)
    node[midway, below=5pt, font=\scriptsize\sffamily, text=red!70!black] {200--2000\,ms};

\node[oatsbox, minimum width=2.2cm, fill=yellow!12, draw=yellow!60!black] (offline) at (4.8, -7.2) {%
    \begin{tabular}{c}
    Offline Learning \\[-2pt]
    {\tiny Stages 1--3}
    \end{tabular}};
\node[cpubox, minimum width=1.6cm, fill=gray!10] (logs) at (1.8, -7.2) {Outcome\\[-1pt]{\tiny Logs}};

\draw[arw, dashed, yellow!60!black]
    (router2.south west) -- ++(0,-0.9) -| (logs.north);
\node[lbl, text=black!60, anchor=south east, fill=white, inner sep=1pt]
    at ([xshift=-0.1cm, yshift=-0.55cm]router2.south west) {\tiny outcomes};

\draw[arw, yellow!60!black] (logs) -- (offline);

\draw[arw, dashed, green!50!black]
    (offline.north) -- ++(0,0.5) -| (router2.south east);
\node[lbl, text=black!60, anchor=south west, fill=white, inner sep=1pt]
    at ([xshift=0.1cm, yshift=-0.55cm]router2.south east) {\tiny updates};

\end{tikzpicture}
\caption{(a)~LLM-based tool selection requires a GPU-bound orchestrator in the request path (500--2,000\,ms). (b)~OATS performs selection on CPU in the router (3--7\,ms), with all learning offline. LLM serving latency is the same in both cases.}
\label{fig:arch_comparison}
\end{figure}

\subsection{Limitations of Static Embedding Similarity}

The standard approach in production routers is \emph{static embedding similarity}: embed each tool's description once, compute dot products against the query embedding at request time, and return the top-$K$. This satisfies the latency constraint but has clear accuracy limitations:

\begin{enumerate}[leftmargin=*]
\item \textbf{Description quality bottleneck.} Tool descriptions are written by developers once and never updated. A poorly worded description permanently under-selects an otherwise useful tool.
\item \textbf{No outcome feedback.} The system cannot learn that Tool~A consistently produces better outcomes than Tool~B for a given query class, even after processing millions of requests.
\item \textbf{Single-point representation.} Each tool is a single embedding vector, which cannot capture tools that serve multiple distinct use cases in different regions of the embedding space.
\item \textbf{Lexical bias.} Embedding similarity rewards surface-level textual similarity between prompts and descriptions rather than functional relevance. Our experiments confirm this: on ToolBench, BM25 (pure lexical) achieves NDCG@5 of 0.853, \emph{exceeding} dense embedding retrieval at 0.834.
\end{enumerate}

These limitations leave room to improve accuracy within the existing latency budget. RL-based tool orchestration~\citep{jin2025searchr1,li2025torl,qian2025toolrl,wang2025otc,su2025toolorchestra} has shown that outcome-aware selection outperforms static methods---but it relies on LLM inference and multi-turn rollouts, which are incompatible with router-level latency constraints.

\subsection{Our Approach}

We propose \textbf{Outcome-Aware Tool Selection (OATS)}. The core idea is straightforward: \emph{interpolate tool embeddings offline toward the centroid of queries where they succeed}. This takes the same idea as RL-based orchestration---learning from outcomes---but implements it as an embedding transformation that adds no parameters, latency, or LLM inference at serving time.

\paragraph{Core mechanism: Embedding-Space Outcome Refinement.} For each tool, OATS collects queries where it was selected and succeeded vs.\ failed, then moves the tool's embedding toward the positive-outcome centroid and away from the negative-outcome centroid. The refined embeddings replace the originals in the tool database as a periodic batch job. At serving time, the path is unchanged: embed the query, compute similarities, return top-$K$. On MetaTool (199 tools), this improves NDCG@5 from 0.869 to 0.940 with no added serving cost. Appendix~\ref{app:examples} walks through real examples where the refinement corrects specific selection errors.

\paragraph{Ablations: learned components.} We also evaluate two extensions to understand when added complexity pays off:
\begin{enumerate}[leftmargin=*]
\item \textbf{Learned Re-Ranking (2,625 parameters).} A small MLP re-scores candidates using outcome-derived features. Adds $<$0.5\,ms.
\item \textbf{Contrastive Embedding Adaptation (197K parameters).} Reshapes the embedding space via hard-negative contrastive learning. Deployed as a drop-in model replacement.
\end{enumerate}

\noindent The MLP re-ranker can actually hurt performance when outcome data is sparse---a result that has direct implications for deployment (Section~\ref{sec:results}).

\subsection{Contributions}

\begin{itemize}[leftmargin=*]
\item An outcome feedback mechanism (Section~\ref{sec:stage1}) that improves tool selection without adding parameters, latency, or GPU cost at serving time.
\item A cost-aware framework (Section~\ref{sec:framework}) that separates the learning and serving concerns, formalizing the latency--accuracy tradeoff.
\item Experiments on ToolBench (2,413 APIs) and MetaTool (Section~\ref{sec:results}) showing when refinement suffices and when learned components add value, including a data-to-tool ratio threshold below which learned re-ranking hurts.
\end{itemize}

\section{Related Work}
\label{sec:related}

\paragraph{Tool-Use in LLMs.}
Tool learning has become a standard way to extend LLM capabilities. Toolformer~\citep{schick2023toolformer} showed that models can learn tool use in a self-supervised manner. ToolLLM~\citep{qin2024toolllm} scaled to 16,000+ real-world APIs and introduced the ToolBench benchmark. MetaTool~\citep{huang2024metatool} benchmarks the \emph{selection} decision---whether to use tools and which to use---and finds that most LLMs still struggle with it.

\paragraph{Tool Retrieval and Reranking.}
As tool sets grow, retrieval becomes necessary before selection. CRAFT~\citep{yuan2024craft} creates task-specific code toolsets and retrieves from them at inference time. ToolRerank~\citep{zheng2024toolrerank} introduces hierarchy-aware reranking that distinguishes between seen and unseen tools and adapts diversity for single- vs.\ multi-tool queries. Our work shares the retrieval perspective but focuses on refining the embedding space using outcome data rather than post-retrieval reranking.

\paragraph{RL for Tool Orchestration.}
Recent work frames tool use as sequential decision-making. Search-R1~\citep{jin2025searchr1} trains LLMs to reason over search results. ToRL~\citep{li2025torl} scales tool-integrated RL; ToolRL~\citep{qian2025toolrl} shows that reward shaping alone can drive tool learning. OTC~\citep{wang2025otc} penalizes excessive tool calls while preserving accuracy, and several studies explore multi-objective rewards balancing correctness, cost, and user preference~\citep{su2025toolorchestra,qian2025toolrl}.

\paragraph{Contrastive Learning for Retrieval.}
The InfoNCE objective~\citep{oord2018cpc} is the basis for most contrastive representation learning. Dense Passage Retrieval~\citep{karpukhin2020dpr} applied contrastive training to retrieval, and Sentence-BERT~\citep{reimers2019sbert} enabled efficient semantic similarity via siamese BERT networks---the backbone we use. R3~\citep{zhou2025r3} applies reinforced contrastive learning to RAG retrieval, generating contrastive signals from environment interactions, which is closely related to our Stage~3.

\paragraph{Semantic Routing.}
Production LLM gateways use semantic routing for model selection, load balancing, and tool attachment. To our knowledge, no prior work has studied outcome-aware tool selection in the context of high-throughput routing proxies, where latency constraints rule out LLM inference for selection. Existing RL-based methods assume the selector and executor share a model or that the selector has its own GPU---neither holds in production routers, which are stateless CPU-bound proxies handling thousands of concurrent requests.

\section{Framework: From Static Matching to Learned Orchestration}
\label{sec:framework}

\subsection{Problem Formulation}

Let $\mathcal{T} = \{t_1, t_2, \ldots, t_n\}$ be a set of available tools, each with a description $d_i \in \mathcal{D}$. Given a user query $q$, the tool selection problem is to choose a subset $S \subseteq \mathcal{T}$ of size at most $K$ that maximizes the expected quality of the downstream LLM response, subject to a \emph{latency budget} $L_{\max}$ and a \emph{resource constraint} $R$:

\begin{equation}
\max_{S \subseteq \mathcal{T}, |S| \leq K} \; \mathbb{E}[\text{quality}(q, S)] \quad \text{s.t.} \quad \text{latency}(S) \leq L_{\max}, \;\; \text{resources}(S) \subseteq R
\label{eq:constrained}
\end{equation}

In production routers, $L_{\max}$ is typically 5--10\,ms and $R$ excludes GPU allocation (which is reserved for model serving). These constraints eliminate any method requiring LLM inference for selection.

\begin{definition}[Tool Selection as Retrieval]
Static tool selection computes:
\begin{equation}
S_{\text{static}}(q) = \operatorname{top\text{-}K}_{t_i \in \mathcal{T}} \; \text{sim}\big(e(q), \, e(d_i)\big)
\label{eq:static}
\end{equation}
where $e(\cdot)$ is an embedding function and $\text{sim}(\cdot, \cdot)$ is cosine similarity or dot product.
\end{definition}

\begin{definition}[Tool Selection as Decision-Making]
RL-based approaches~\citep{jin2025searchr1,li2025torl,su2025toolorchestra} formulate tool selection as an MDP $\mathcal{M} = (\mathcal{U}, \mathcal{S}, \mathcal{A}, \mathcal{O}, \mathcal{T}, \mathcal{Z}, r, \rho, \gamma)$ where an orchestrator policy $\pi_\theta(a_k | h_k)$ selects tools sequentially with a multi-objective reward:
\begin{equation}
R(\tau) = M_{\text{normalized}}^\tau \cdot P
\label{eq:rl_reward}
\end{equation}
where $M^\tau$ encodes tool usage counts, outcome, cost, and latency, and $P$ is a user preference vector.
\end{definition}

\subsection{Bridging the Gap: Offline Learning, Online Retrieval}

OATS approximates the outcome-aware behavior of Equation~\ref{eq:rl_reward} within the constraints of Equations~\ref{eq:static} and~\ref{eq:constrained}. The design principle is to separate learning from serving: all outcome-aware optimization happens offline from logged data, while the serving path remains a fast embedding lookup. We write the resulting scoring function as:

\begin{equation}
S_{\text{OATS}}(q) = \operatorname{top\text{-}K}_{t_i \in \mathcal{T}} \; f_\phi\big(e_\psi(q), \, e_\psi(d_i), \, m_i, \, h_i(q)\big)
\label{eq:oats}
\end{equation}

where $f_\phi$ is a learned scoring function, $e_\psi$ are (potentially fine-tuned) embeddings, $m_i$ is tool metadata, and $h_i(q)$ encodes historical outcome statistics for tool $t_i$ on queries similar to $q$. The three OATS stages correspond to optimizing different components of Equation~\ref{eq:oats}:

\begin{itemize}[leftmargin=*]
\item \textbf{Stage~1} replaces $e(d_i)$ (the stored tool embeddings) via offline centroid interpolation, keeping $f$ and $h$ fixed.
\item \textbf{Stage~2} learns $f_\phi$ while keeping $e$ and $d_i$ fixed.
\item \textbf{Stage~3} jointly optimizes $e_\psi$ via contrastive learning.
\end{itemize}

\subsection{Connection to RL Reward Normalization}

Policy-gradient methods such as GRPO~\citep{shao2024deepseekmath} compute a normalized advantage over a batch of rollouts:
\begin{equation}
A(\tau) = \frac{R(\tau) - \text{mean}_{\tau \in T} R(\tau)}{\text{std}_{\tau \in T} R(\tau)}
\end{equation}
This creates a contrastive signal: trajectories with better tool selections receive positive advantage relative to the batch. Our Stage~3 contrastive loss implements the same idea in embedding space:
\begin{equation}
\mathcal{L}_{\text{contrastive}} = -\log \frac{\exp(\text{sim}(e(q), e(d^+)) / \tau)}{\sum_{d^- \in \mathcal{N}(q)} \exp(\text{sim}(e(q), e(d^-)) / \tau)}
\label{eq:contrastive}
\end{equation}
where $d^+$ is a tool description that led to a positive outcome for query $q$, and $\mathcal{N}(q)$ contains descriptions of tools that led to negative outcomes. This mirrors policy-gradient advantage computation but operates in embedding space rather than over policy parameters, without the cost of trajectory rollouts.

\section{Method}
\label{sec:method}

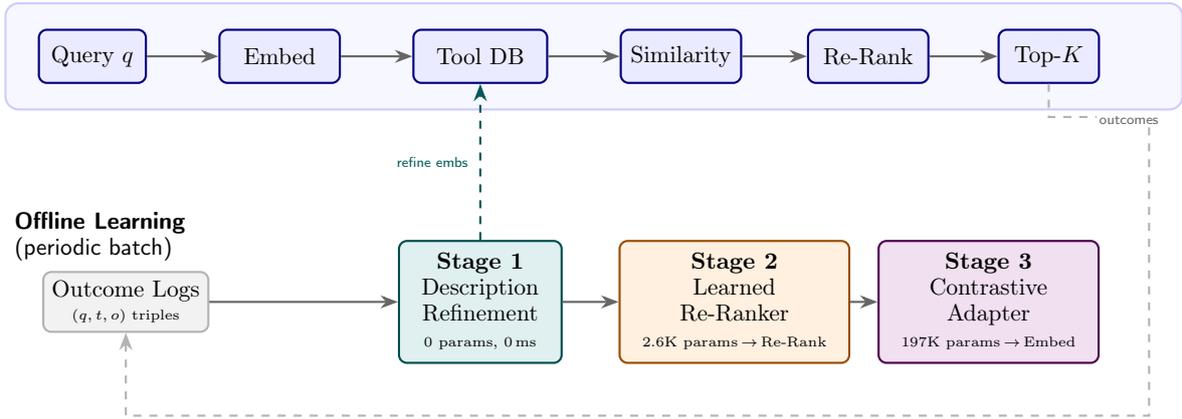
\begin{figure*}[t]
\centering
\begin{tikzpicture}[
    >=Stealth,
    every node/.style={font=\small},
    box/.style={draw, rounded corners=3pt, minimum height=0.8cm, align=center, thick},
    stage/.style={box, minimum width=2.0cm, minimum height=1.0cm},
    s1/.style={stage, fill=teal!12, draw=teal!60!black},
    s2/.style={stage, fill=orange!12, draw=orange!60!black},
    s3/.style={stage, fill=violet!12, draw=violet!60!black},
    pipenode/.style={box, fill=blue!8, draw=blue!50!black, minimum width=1.6cm},
    datanode/.style={box, fill=gray!10, draw=gray!60, minimum width=1.5cm},
    arw/.style={->, thick, color=black!60},
    darw/.style={->, thick, dashed, color=black!40},
    lbl/.style={font=\scriptsize\sffamily, text=black!60},
    stagelab/.style={font=\scriptsize\sffamily},
    scale=0.88, transform shape,
]

\node[font=\small\bfseries\sffamily, anchor=west] at (-0.3, 2.5) {Online Serving Path {\normalfont\small(per request, CPU, 3--7\,ms)}};

\draw[blue!20, thick, rounded corners=5pt, fill=blue!3]
    (-0.3, 2.1) rectangle (17.3, 0.5);

\node[pipenode] (query) at (1.0, 1.3) {Query $q$};
\node[pipenode, minimum width=1.8cm] (embed) at (3.8, 1.3) {Embed};
\node[pipenode, minimum width=2.0cm] (tooldb) at (6.8, 1.3) {Tool DB};
\node[pipenode, minimum width=1.8cm] (sim) at (9.8, 1.3) {Similarity};
\node[pipenode, minimum width=1.8cm] (rerank) at (12.6, 1.3) {Re-Rank};
\node[pipenode, minimum width=1.5cm] (topk) at (15.3, 1.3) {Top-$K$};

\draw[arw] (query) -- (embed);
\draw[arw] (embed) -- (tooldb);
\draw[arw] (tooldb) -- (sim);
\draw[arw] (sim) -- (rerank);
\draw[arw] (rerank) -- (topk);


\node[datanode, minimum width=2.4cm] (logs) at (1.5, -2.4) {Outcome Logs\\[-2pt]{\tiny $(q, t, o)$ triples}};

\node[font=\small\bfseries\sffamily, anchor=west] at (-0.3, -1.2) {Offline Learning};
\node[font=\small\sffamily, anchor=west] at (-0.3, -1.6) {(periodic batch)};

\node[s1] (st1) at (6.8, -2.4) {%
    \begin{tabular}{c}
    \textbf{Stage 1} \\[-1pt]
    Description \\[-1pt]
    Refinement \\[-1pt]
    {\tiny 0 params, 0\,ms}
    \end{tabular}};

\node[s2] (st2) at (10.6, -2.4) {%
    \begin{tabular}{c}
    \textbf{Stage 2} \\[-1pt]
    Learned \\[-1pt]
    Re-Ranker \\[-1pt]
    {\tiny 2.6K params\,$\to$\,Re-Rank}
    \end{tabular}};

\node[s3] (st3) at (14.4, -2.4) {%
    \begin{tabular}{c}
    \textbf{Stage 3} \\[-1pt]
    Contrastive \\[-1pt]
    Adapter \\[-1pt]
    {\tiny 197K params\,$\to$\,Embed}
    \end{tabular}};

\draw[arw] (logs) -- (st1);
\draw[arw] (st1) -- (st2);
\draw[arw] (st2) -- (st3);

\draw[darw, teal!70!black, thick]
    (st1.north) -- (tooldb.south);
\node[stagelab, text=teal!70!black, anchor=east, fill=white, inner sep=2pt]
    at ([xshift=-3pt]$(st1.north)!0.5!(tooldb.south)$) {\tiny refine embs};

\draw[darw, gray!60]
    (topk.south) -- ++(0,-0.5) -- ++(1.5,0) -- ++(0,-4.5) -| (logs.south);
\node[lbl, fill=white, inner sep=1pt]
    at ([xshift=1.2cm, yshift=-0.55cm]topk.south) {\tiny outcomes};

\end{tikzpicture}
\caption{The OATS pipeline. \textbf{Top:} the online serving path runs on CPU in 3--7\,ms. \textbf{Bottom:} offline learning from outcome logs. The core mechanism (Stage~1, dashed arrow) refines tool embeddings in the Tool DB at zero serving cost. Stages~2 and~3 are ablation mechanisms that optionally update the Re-Rank and Embed components respectively.}
\label{fig:oats_pipeline}
\end{figure*}

We first describe the core OATS mechanism---embedding-space outcome refinement---and then two learned extensions (re-ranking and contrastive adaptation) evaluated as ablations. Figure~\ref{fig:oats_pipeline} gives an overview.

\subsection{Core Mechanism: Outcome-Guided Embedding Refinement}
\label{sec:stage1}

\paragraph{Resource profile.} No inference cost, no additional parameters at serving time. The only requirement is a periodic offline batch job (hourly or daily).

\paragraph{What it does and does not change.} OATS-S1 does \emph{not} rewrite or modify tool descriptions. The text stays the same. What changes is the \emph{embedding vector} stored in the tool database: the original vector $e(d_i)$---computed from the description---is replaced by a refined vector that better reflects the tool's actual usage, as learned from outcome data. At serving time, the selection path is identical to the static baseline (embed query, compute similarities, return top-$K$); only the stored tool vectors differ.

\paragraph{Motivating example.} Consider a tool called \texttt{buildbetter} whose description reads: ``Chat with the knowledge of all your calls in BuildBetter (Zoom, GMeet, Webex). Start for free @ BuildBetter.ai.'' This is a meeting-transcript tool, but the description is a marketing tagline. When a user asks ``\emph{provide the complete transcript of the strategy call with the executives},'' a static embedding model ranks the financial-data tool \texttt{QuiverQuantitative} first (cosine similarity 0.337) and the correct tool second (0.276), because ``strategy\ldots executives'' is closer to ``congressional\ldots lobbying\ldots legislation'' in embedding space than to ``BuildBetter.ai.''

\begin{figure}[t]
\centering
\begin{tikzpicture}[
    >=Stealth,
    scale=0.88, transform shape,
    every node/.style={font=\small},
    qpos/.style={circle, inner sep=1.8pt, fill=teal!70, draw=teal!50!black, line width=0.4pt},
    qneg/.style={circle, inner sep=1.8pt, fill=red!60, draw=red!40!black, line width=0.4pt},
    cpos/.style={regular polygon, regular polygon sides=4, inner sep=2.5pt, fill=teal!60, draw=teal!40!black, line width=0.6pt, rotate=45},
    cneg/.style={regular polygon, regular polygon sides=4, inner sep=2.5pt, fill=red!50, draw=red!40!black, line width=0.6pt, rotate=45},
    toolorig/.style={circle, inner sep=3pt, fill=blue!50, draw=blue!40!black, line width=0.8pt},
    toolref/.style={circle, inner sep=3pt, fill=teal!70, draw=teal!50!black, line width=0.8pt},
    toolother/.style={circle, inner sep=3pt, fill=gray!50, draw=gray!40!black, line width=0.8pt},
    lbl/.style={font=\scriptsize},
]

\fill[blue!4, rounded corners=12pt] (-1.8,-2.5) rectangle (2.5,1.2);
\node[font=\tiny\sffamily, text=blue!30, anchor=north west] at (-1.7,1.1) {SaaS / productivity};

\fill[teal!6, rounded corners=12pt] (3.0,-0.3) rectangle (8.5,3.8);
\node[font=\tiny\sffamily, text=teal!35, anchor=north west] at (3.1,3.7) {meeting transcripts / calls};

\fill[red!5, rounded corners=12pt] (-0.5,-4.5) rectangle (4.5,-1.8);
\node[font=\tiny\sffamily, text=red!30, anchor=south west] at (-0.4,-4.4) {call management (CLINQ)};

\node[qpos] (qp1) at (4.2, 2.8) {};
\node[qpos] (qp2) at (5.6, 2.2) {};
\node[qpos] (qp3) at (4.8, 1.4) {};
\node[qpos] (qp4) at (6.2, 3.0) {};

\node[lbl, text=teal!50!black, anchor=west] at (6.5,3.0) {``transcript of call''};
\node[lbl, text=teal!50!black, anchor=west] at (5.9,2.2) {``key points from meeting''};

\node[cpos] (centpos) at (5.2, 2.35) {};
\node[lbl, text=teal!50!black, anchor=south, yshift=2pt] at (centpos.north) {$\bar{e}^+$};

\node[qneg] (qn1) at (1.2, -3.2) {};
\node[qneg] (qn2) at (2.5, -2.6) {};
\node[qneg] (qn3) at (3.2, -3.6) {};

\node[lbl, text=red!50!black, anchor=north] at (2.5,-3.8) {``create call reminder on CLINQ''};

\node[cneg] (centneg) at (2.3, -3.1) {};
\node[lbl, text=red!50!black, anchor=north east, xshift=-2pt] at (centneg.south) {$\bar{e}^-$};

\node[toolorig] (orig) at (0.8, -0.3) {};
\node[lbl, anchor=east, xshift=-4pt, text=blue!40!black] at (orig.west) {%
  \begin{tabular}{r}
  \texttt{buildbetter}\\[-1pt]
  {\tiny original $e(d_i)$}
  \end{tabular}};

\node[toolref] (refined) at (3.6, 1.0) {};
\node[lbl, anchor=west, xshift=4pt, text=teal!50!black] at (refined.east) {%
  \begin{tabular}{l}
  \texttt{buildbetter}\\[-1pt]
  {\tiny refined $\hat{e}_i$}
  \end{tabular}};

\node[star, star points=5, star point ratio=2.2, inner sep=1.5pt, fill=orange!70, draw=orange!50!black, line width=0.5pt] (testq) at (5.0, 0.6) {};
\node[lbl, text=orange!50!black, anchor=west, xshift=3pt] at (testq.east) {test query};

\node[toolother] (quiver) at (1.5, 0.8) {};
\node[lbl, anchor=south, yshift=2pt, text=gray!50!black] at (quiver.north) {\texttt{QuiverQuant.}};

\draw[->, thick, teal!60!black, line width=1.2pt]
    (orig) -- node[lbl, above, sloped, text=teal!50!black, yshift=1pt] {$+\alpha \cdot \bar{e}^+$} ($(orig)!0.55!(centpos)$);

\draw[->, thick, red!50!black, line width=1.0pt, dashed]
    (orig) -- node[lbl, below, sloped, text=red!40!black, yshift=-1pt] {$-\beta \cdot \bar{e}^-$} ($(orig)!0.35!(centneg)$);

\draw[->, thick, black!40, line width=0.8pt, dotted]
    (orig) to[bend left=15] (refined);

\node[qpos, scale=0.8] at (-1.5, -5.1) {};
\node[lbl, anchor=west] at (-1.2, -5.1) {$Q^+$ queries};
\node[qneg, scale=0.8] at (1.0, -5.1) {};
\node[lbl, anchor=west] at (1.3, -5.1) {$Q^-$ queries};
\node[star, star points=5, star point ratio=2.2, inner sep=1pt, fill=orange!70, draw=orange!50!black, line width=0.4pt, scale=0.8] at (3.5, -5.1) {};
\node[lbl, anchor=west] at (3.8, -5.1) {test query};

\node[cpos, scale=0.7] at (-1.5, -5.6) {};
\node[lbl, anchor=west] at (-1.2, -5.6) {pos.\ centroid};
\node[cneg, scale=0.7] at (1.0, -5.6) {};
\node[lbl, anchor=west] at (1.3, -5.6) {neg.\ centroid};
\node[toolorig, scale=0.6] at (3.3, -5.6) {};
\node[lbl, anchor=center] at (3.6, -5.6) {/};
\node[toolref, scale=0.6] at (3.9, -5.6) {};
\node[lbl, anchor=west] at (4.2, -5.6) {orig.\ / refined};

\end{tikzpicture}
\caption{Geometry of OATS-S1 for the \texttt{buildbetter} example. The original embedding (blue circle) sits in a generic ``SaaS'' region, far from the test query (star). Positive training queries $Q^+$ (teal dots) cluster around ``meeting transcripts''; negative queries $Q^-$ (red dots) cluster around ``call management.'' The refinement pulls the tool embedding toward $\bar{e}^+$ and away from $\bar{e}^-$, placing the refined embedding (teal circle) closer to the test query. The description text never changes.}
\label{fig:s1_geometry}
\end{figure}
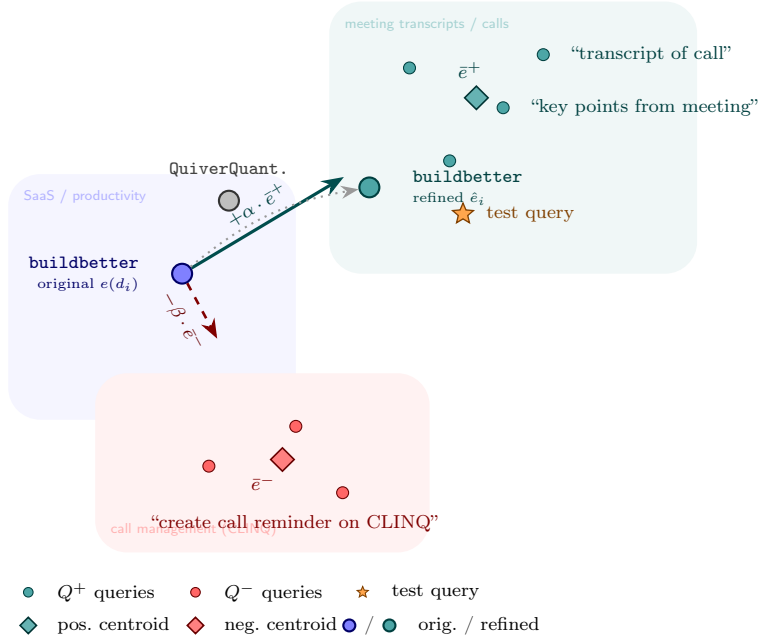

After S1 refinement, the \texttt{buildbetter} embedding has been pulled toward the centroid of queries like ``\emph{retrieve the transcript of the customer support call}'' and ``\emph{what were the key points from last week's meeting}''---queries where it was the correct tool in the training set. Its similarity to the test query jumps from 0.276 to 0.440, while \texttt{QuiverQuantitative} barely moves (0.337 $\to$ 0.343). The description text is unchanged; only the vector in the database differs. Figure~\ref{fig:s1_geometry} illustrates the geometry; Appendix~\ref{app:examples} gives the full walkthrough.

\paragraph{Algorithm.} The procedure is in Algorithm~\ref{alg:s1}. We explain each step below.

\begin{algorithm}[t]
\caption{OATS-S1: Iterative Outcome-Guided Embedding Refinement}
\label{alg:s1}
\begin{algorithmic}[1]
\REQUIRE Tool set $\mathcal{T}$ with embeddings $\{e(d_i)\}$, training queries with ground-truth labels, iterations $N$, step sizes $\alpha$, $\beta$, momentum $\mu = 0.5$
\ENSURE Refined embeddings $\{e'(d_i)\}$ for each tool
\STATE Initialize $e^{(0)}(d_i) \leftarrow e(d_i)$ for all $t_i \in \mathcal{T}$
\FOR{$n = 1$ \TO $N$}
    \STATE \textbf{// Step 1: Build outcome logs using current embeddings}
    \FOR{each training query $q_j$}
        \STATE Retrieve top-$K$ tools by $\text{sim}(e(q_j),\, e^{(n-1)}(d_i))$
        \STATE Label each retrieved tool: $o = 1$ if $t_i \in \text{ground\_truth}(q_j)$, else $o = 0$
    \ENDFOR
    \STATE \textbf{// Step 2: Partition queries by tool and outcome}
    \FOR{each tool $t_i$}
        \STATE $Q_i^+ \leftarrow \{q_j : t_i \text{ retrieved for } q_j \text{ and } o_j = 1\}$
        \STATE $Q_i^- \leftarrow \{q_j : t_i \text{ retrieved for } q_j \text{ and } o_j = 0\}$
    \ENDFOR
    \STATE \textbf{// Step 3: Compute refined embedding per tool}
    \FOR{each tool $t_i$ with $|Q_i^+| \geq 1$}
        \STATE $\bar{e}^+ \leftarrow \frac{1}{|Q_i^+|}\sum_{q \in Q_i^+} e(q)$ \hfill (positive centroid)
        \STATE $\hat{e}_i \leftarrow (1 - \alpha)\, e^{(n-1)}(d_i) \;+\; \alpha\, \bar{e}^+$
        \IF{$|Q_i^-| \geq 1$}
            \STATE $\bar{e}^- \leftarrow \frac{1}{|Q_i^-|}\sum_{q \in Q_i^-} e(q)$ \hfill (negative centroid)
            \STATE $\hat{e}_i \leftarrow \hat{e}_i \;-\; \beta\, \bar{e}^-$ \hfill (repel from failures)
        \ENDIF
        \STATE $\hat{e}_i \leftarrow \hat{e}_i \,/\, \|\hat{e}_i\|$ \hfill (re-normalize)
        \STATE \textbf{// Step 4: Momentum blend with previous iteration}
        \IF{$n > 1$}
            \STATE $e^{(n)}(d_i) \leftarrow \mu\, e^{(n-1)}(d_i) + (1 - \mu)\, \hat{e}_i$
            \STATE $e^{(n)}(d_i) \leftarrow e^{(n)}(d_i) \,/\, \|e^{(n)}(d_i)\|$
        \ELSE
            \STATE $e^{(n)}(d_i) \leftarrow \hat{e}_i$
        \ENDIF
    \ENDFOR
\ENDFOR
\STATE \textbf{// Step 5: Validation gate}
\STATE Evaluate $\{e^{(N)}(d_i)\}$ on held-out validation set
\STATE Accept $e'(d_i) \leftarrow e^{(N)}(d_i)$ only if Recall@$K$ improves over $e(d_i)$
\end{algorithmic}
\end{algorithm}

\paragraph{Step 1--2: Outcome collection and partitioning.} From production logs (or ground-truth labels in benchmarks), we collect tuples $(q_j, t_i, o_j)$ where $q_j$ is a query, $t_i$ a retrieved tool, and $o_j \in \{0, 1\}$ indicates whether $t_i$ was relevant. For each tool $t_i$, this gives a positive set $Q_i^+$ (queries where $t_i$ was correct) and a negative set $Q_i^-$ (queries where $t_i$ was retrieved but wrong---hard negatives).

Because $Q_i^-$ contains only high-similarity negatives rather than random ones, the repulsion signal is targeted: the tool is pushed away specifically from the queries that currently cause false matches.

\paragraph{Step 3: Centroid interpolation.} Each tool embedding is moved toward its positive centroid and away from its negative centroid:
\begin{equation}
\hat{e}_i = (1 - \alpha) \cdot e(d_i) + \alpha \cdot \bar{e}(Q_i^+) - \beta \cdot \bar{e}(Q_i^-)
\label{eq:desc_refine}
\end{equation}
where $\bar{e}(Q_i^+) = \frac{1}{|Q_i^+|}\sum_{q \in Q_i^+} e(q)$ is the positive centroid, $\bar{e}(Q_i^-)$ is the negative centroid, $\alpha = 0.3$ controls the attraction strength, and $\beta = 0.1$ controls the repulsion strength ($\beta < \alpha$ because false negatives in $Q_i^-$ are more common than false positives in $Q_i^+$). The result is re-normalized to unit length.

\paragraph{Step 4: Iterative refinement with momentum.} A single pass may not fully correct embeddings, because the outcome logs depend on the current rankings. After iteration 1, the top-$K$ retrievals change, exposing new hard negatives. We iterate $N = 3$ times, re-computing outcomes against updated embeddings at each step. To prevent oscillation, each iteration blends with the previous estimate via momentum $\mu = 0.5$: $e^{(n)} = \mu \cdot e^{(n-1)} + (1-\mu) \cdot \hat{e}$.

\paragraph{Step 5: Validation gate.} The refined embeddings are accepted only if they improve Recall@$K$ on a held-out validation set. This prevents drift from noisy outcomes---the system cannot degrade below the static baseline. In our experiments (Section~\ref{sec:results}), the gate always accepts.

\paragraph{Why it works.} Appendix~\ref{app:examples} gives four worked examples with real similarity scores. The refinement helps with three failure modes of static embeddings: (1)~semantic decoys---tools with superficially similar descriptions but different functions; (2)~opaque descriptions---branded or generic names that do not embed well; and (3)~low-similarity regimes---when candidates score similarly and small interpolations flip rankings.

\subsection{Ablation Mechanism A: Learned Re-Ranking}
\label{sec:stage2}

We next ask whether a learned component on top of refined embeddings improves selection, evaluating a lightweight MLP re-ranker as the simplest learned intervention.

\paragraph{Resource profile.} 2,625 trainable parameters (architecture: $[7, 64, 32, 1]$). Adds $<$0.5\,ms p50 latency. Runs entirely on CPU. The model binary is $\sim$11\,KB.

\paragraph{Architecture.} We train a lightweight MLP $f_\phi : \mathbb{R}^{d_{\text{feat}}} \to [0,1]$ on features derived from the candidate set:
\begin{equation}
\text{features}(q, t_i) = \big[\text{sim}(e(q), e(d_i)), \; \Delta_{\text{sim}}, \; \text{cat}(t_i), \; \text{sr}_i(q), \; \text{freq}_i, \; \text{len}(q)\big]
\end{equation}

where $\Delta_{\text{sim}}$ is the similarity gap to the next candidate, $\text{cat}(t_i)$ is a category indicator, $\text{sr}_i(q)$ is the historical success rate of tool $t_i$ on queries in the same cluster as $q$, $\text{freq}_i$ is tool usage frequency, and $\text{len}(q)$ is the query length.

The re-ranker is trained with binary cross-entropy:
\begin{equation}
\mathcal{L}_{\text{rerank}} = -\frac{1}{N}\sum_{j=1}^{N} \big[o_j \log f_\phi(\text{features}(q_j, t_j)) + (1-o_j)\log(1-f_\phi(\text{features}(q_j, t_j)))\big]
\end{equation}

At inference time, we retrieve $C = \alpha K$ candidates via static similarity ($\alpha = 5$), then re-rank with $f_\phi$ and return the top-$K$. The re-ranker needs enough training data per tool to be useful---roughly 10 outcome examples per tool as a minimum (Section~\ref{sec:results}).

\subsection{Ablation Mechanism B: Contrastive Embedding Adaptation}
\label{sec:stage3}

The second ablation reshapes the embedding space directly via contrastive learning, rather than adding a post-hoc re-ranker. This may scale better to large tool sets where per-tool feature learning is data-starved.

\paragraph{Resource profile.} 197K adapter parameters (two-layer projection head). At serving time, the adapter runs a single matrix multiply on the 384-dimensional embedding, adding $<$0.1\,ms. The adapter can be fused into the embedding model weights for zero overhead. Total model size increase: $<$1\,MB.

\paragraph{Triplet mining.} From outcome logs, we mine triplets $(q, d^+, d^-)$ where $d^+$ is the description of a tool that produced a positive outcome for $q$, and $d^-$ is a \emph{hard negative}---a tool with high embedding similarity to $q$ but poor outcomes. Hard negatives are critical for learning the functional boundaries that static embeddings miss.

\paragraph{Training objective.} We train the adapter with InfoNCE loss~\citep{oord2018cpc} (Equation~\ref{eq:contrastive}), combining in-batch negatives with mined hard negatives. We use a small learning rate ($10^{-5}$) and early stopping on validation NDCG to avoid degrading general embedding quality. We train an adapter head rather than fine-tuning the full model, which reduces training cost, preserves base model quality, and allows instant rollback by disabling the adapter.

\paragraph{Deployment.} The adapter is placed alongside the embedding model at the router level. Since the output dimension is unchanged, it is a drop-in replacement---\texttt{ToolsDatabase}, similarity computation, and \texttt{FilterAndRankTools} all remain the same. Tool embeddings are recomputed once after deployment ($<$2 seconds for 2,413 tools).

\section{Experimental Setup}
\label{sec:experiments}

\subsection{Datasets}

\paragraph{ToolBench~\citep{qin2024toolllm}.} We use the benchmark split from the ToolBench dataset, extracting 2,413 unique APIs across 46 categories and 600 queries across three evaluation settings of increasing difficulty: G1-Instruction (single-tool, 200 queries), G1-Category (intra-category, 200 queries), and G2-Instruction (multi-tool, 200 queries). We treat the annotated \texttt{relevant\_apis} as ground truth for tool selection evaluation.

\paragraph{MetaTool~\citep{huang2024metatool}.} We use the Task~2 (tool selection) data from the official MetaTool repository, which contains 4,287 queries across 199 tools with four subtask types: similar choices (995 queries), specific scenarios (1,800 queries), reliability issues (995 queries), and multi-tool selection (497 queries). Each query includes $\sim$10 candidate tools and human-annotated ground-truth tool(s).

\subsection{Evaluation Metrics}

We report standard retrieval metrics:
\begin{itemize}[leftmargin=*]
\item \textbf{Recall@$K$}: Fraction of ground-truth tools present in the top-$K$ selected tools.
\item \textbf{Precision@$K$}: Fraction of selected tools that are ground-truth relevant.
\item \textbf{NDCG@$K$}: Normalized Discounted Cumulative Gain measuring ranking quality.
\item \textbf{MRR}: Mean Reciprocal Rank of the first relevant tool.
\end{itemize}

We report results for $K \in \{1, 3, 5\}$ to capture different operational settings.

\subsection{Baselines}

\begin{enumerate}[leftmargin=*]
\item \textbf{BM25}: Sparse lexical retrieval over tool descriptions.
\item \textbf{Static Embedding (SE)}: Dense retrieval using sentence-transformers with original tool descriptions (analogous to the current semantic router).
\item \textbf{SE + Lexical}: Static embedding with the lexical/tag/name/category weighted combination from the semantic router's \texttt{FilterAndRankTools}.
\item \textbf{Random}: Random tool selection (lower bound).
\end{enumerate}

\subsection{OATS Configurations}

\begin{enumerate}[leftmargin=*]
\item \textbf{OATS-S1}: Core mechanism only (embedding refinement via outcome-guided centroid interpolation).
\item \textbf{OATS-S2}: Core + ablation A (embedding refinement + MLP re-ranker).
\item \textbf{OATS-S3}: Core + ablation A + ablation B (embedding refinement + MLP re-ranker + contrastive adapter).
\end{enumerate}

\subsection{Implementation Details}

The base embedding model is \texttt{all-MiniLM-L6-v2}~\citep{reimers2019sbert} (22M parameters, 384 dimensions), representative of models deployed in production routers. The re-ranker MLP is $[7, 64, 32, 1]$ (2,625 parameters, ReLU, dropout 0.1). The contrastive adapter is $[384, 256, 384]$ (197K parameters), trained with learning rate $10^{-5}$, temperature $\tau = 0.07$, up to 5 epochs with early stopping. All experiments use $\alpha = 5$ for the re-ranking candidate pool.

\paragraph{Outcome labels.} Both benchmarks provide human-annotated tool relevance per query. We use these as outcome labels: $o_j = 1$ if tool $t_j$ is in the annotated relevant set for $q_j$, and $o_j = 0$ otherwise. In production, $o_j$ would come from downstream signals (task completion, user satisfaction); the framework is agnostic to the source.

\paragraph{Train/test protocol.} All methods are evaluated on the same held-out 30\% test split (fixed 70/30 split, deterministic seed). OATS stages use the 70\% training portion for learning; Stage~2 further sub-splits training into 85/15 train/validation.

\paragraph{Latency measurement.} All latencies are measured on a single CPU core (no GPU), covering embedding computation, similarity search, and any re-ranking overhead. We report p50 and p99.

\section{Results}
\label{sec:results}

We evaluate each method on selection accuracy and serving cost.

\subsection{Core Result: Zero-Cost Refinement}

\paragraph{MetaTool (199 tools, 1,287 test queries).}
This benchmark covers four subtask types including adversarially similar tool choices and multi-tool selection. Static embedding achieves Recall@1 = 0.716 and NDCG@5 = 0.869. OATS embedding refinement improves this to Recall@1 = 0.830 and NDCG@5 = 0.940, with no added serving cost (p50: 3.53\,ms vs.\ 3.66\,ms). Appendix~\ref{app:examples} shows concrete before/after examples.

\paragraph{ToolBench (2,413 tools, 180 test queries).}
With a larger tool set, all methods are stressed. BM25 (pure lexical) achieves NDCG@5 of 0.853, exceeding static embedding at 0.834---suggesting that ToolBench queries have enough lexical overlap with API descriptions for sparse retrieval to work well. OATS embedding refinement reaches NDCG@5 of 0.848 (+1.7\% over dense embedding) at slightly lower latency (p50: 4.42\,ms vs.\ 4.59\,ms). The more modest gain compared to MetaTool is expected: with 2,413 tools, the embedding space is densely packed and centroid interpolation has less room to separate tools without affecting neighbors.

\subsection{Ablation: Do Learned Components Help Further?}

Table~\ref{tab:ablation} summarizes the incremental contribution and cost of each learned component on top of refined embeddings.

\paragraph{MLP re-ranker (2.6K parameters).}
The re-ranker hurts on ToolBench (NDCG@5: 0.823 vs.\ 0.834 baseline) and is flat on MetaTool (0.869 vs.\ 0.869). On ToolBench, 357 training queries spread across 2,413 tools gives $< 0.15$ positive examples per tool. Even on MetaTool ($\sim$13 examples/tool), the MLP adds nothing. Either the feature set (similarity gap, category indicator, etc.) lacks discriminative power for this task, or the MLP cannot generalize from sparse per-tool statistics. In short, a learned re-ranker is not guaranteed to help and should be validated before deployment.

\paragraph{Contrastive adapter (197K parameters).}
The adapter reaches NDCG@5 of 0.842 on ToolBench (+0.008 over baseline) and 0.931 on MetaTool (+0.062). On MetaTool the adapter performs comparably to embedding refinement (0.931 vs.\ 0.940), suggesting both capture similar outcome signals at this scale. Unlike the MLP, the adapter reshapes the full embedding space via hard-negative contrastive learning rather than learning per-tool features.

\subsection{Latency and Resource Analysis}

Table~\ref{tab:latency} shows that all OATS stages stay within single-digit millisecond p50 latency on both benchmarks; Stage~1 is comparable to or faster than the baseline.

\paragraph{Cost-efficiency metric.} We define accuracy gain per millisecond (AG/ms) as $\Delta\text{NDCG@5} / \Delta\text{latency}$. When accuracy improves with no latency increase, we assign $\infty$. Table~\ref{tab:efficiency} summarizes.

\begin{table}[h]
\centering
\caption{Cost efficiency: NDCG@5 gain per additional millisecond of p50 latency vs.\ SE baseline. $\infty$ means accuracy improved with no latency increase.}
\label{tab:efficiency}
\begin{tabular}{l cc cc}
\toprule
& \multicolumn{2}{c}{\textbf{MetaTool}} & \multicolumn{2}{c}{\textbf{ToolBench}} \\
\cmidrule(lr){2-3} \cmidrule(lr){4-5}
\textbf{Method} & $\Delta$NDCG@5 & AG/ms & $\Delta$NDCG@5 & AG/ms \\
\midrule
SE (baseline)               & ---    & ---       & ---    & ---       \\
Emb.\ refinement (core)    & +0.071 & $\infty$  & +0.014 & $\infty$  \\
+ Contrastive (ablation B)  & +0.062 & $\infty$  & +0.008 & $\infty$  \\
SE + Lexical                & -0.053 & ---       & +0.020 & $\infty$  \\
\bottomrule
\end{tabular}
\end{table}

\noindent Embedding refinement improves accuracy at no latency cost on both benchmarks. On ToolBench, SE + Lexical achieves the best accuracy (0.854), but OATS refinement is close (0.848) without lexical features.

\subsection{Key Findings}

Four findings for practitioners:

\begin{enumerate}[leftmargin=*]
\item \textbf{Start with embedding refinement.} It is free at inference time, needs no code changes, and gives the largest gains on MetaTool (NDCG@5 0.869\,$\to$\,0.940) with meaningful improvement on ToolBench (0.834\,$\to$\,0.848).

\item \textbf{An MLP re-ranker does not reliably help.} It fails to improve on either benchmark and hurts on ToolBench. Do not assume a learned re-ranker will help without validation on your data.

\item \textbf{Contrastive adaptation is an alternative.} On MetaTool it reaches NDCG@5 0.931 vs.\ 0.940 for refinement. Useful for teams that cannot modify embedding tables but can swap models.

\item \textbf{Lexical features remain strong on ToolBench.} BM25 beats dense embeddings, and SE + Lexical gets the best NDCG@5 (0.854). Refinement should be evaluated on top of lexical features in production.
\end{enumerate}

\subsection{Comparison with LLM-Based Tool Selection}

Table~\ref{tab:llm_comparison} compiles published LLM-based results on MetaTool~\citep{huang2024metatool} alongside OATS.

\begin{table}[h]
\centering
\caption{Comparison with LLM-based tool selection on the MetaTool ``similar choices'' subtask (295 test queries). LLM results report Correct Selection Rate (CSR) from~\citet{huang2024metatool}; retrieval-based results report Recall@1 on the same subtask and split. Both metrics measure whether the top-selected tool is correct.}
\label{tab:llm_comparison}
\begin{tabular}{l c r l}
\toprule
\textbf{Method} & \textbf{Accuracy} & \textbf{Latency} & \textbf{Hardware} \\
\midrule
\multicolumn{4}{l}{\emph{LLM-based (from published benchmarks)}} \\
\quad ChatGPT (GPT-3.5)$^\dagger$       & 69.1\%  & $\sim$1--3\,s & API / GPU \\
\quad Vicuna-7b$^\dagger$                & 73.5\%  & $\sim$2--5\,s & GPU \\
\quad Vicuna-13b$^\dagger$               & 58.2\%  & $\sim$3--8\,s & GPU \\
\quad LLaMA2-13b$^\dagger$              & 44.1\%  & $\sim$2--5\,s & GPU \\
\quad Average (9 LLMs)$^\dagger$         & $\sim$57\%   & ---  & GPU \\
\midrule
\multicolumn{4}{l}{\emph{Retrieval-based (ours, same subtask, CPU-only)}} \\
\quad BM25                           & 42.7\%  & 0.4\,ms & CPU \\
\quad Static Embedding (SE)          & 66.4\%  & 3.7\,ms & CPU \\
\quad \textbf{OATS-S1 (emb.\ refine)}  & \textbf{83.4\%}  & \textbf{3.5\,ms} & \textbf{CPU} \\
\bottomrule
\multicolumn{4}{l}{\footnotesize $^\dagger$\,CSR on ``tool selection with similar choices'' subtask~\citep{huang2024metatool}.} \\
\multicolumn{4}{l}{\footnotesize LLM latencies are approximate wall-clock times for full inference.}
\end{tabular}
\end{table}

\noindent On the ``similar choices'' subtask---the hardest split where tools share overlapping functionality---OATS-S1 reaches 83.4\%, roughly 10 points above Vicuna-7b (73.5\%) and 14 points above ChatGPT (69.1\%), while running $\sim$1,000$\times$ faster on CPU without any language-model inference.

Caveats apply: (1)~MetaTool reports CSR (correct selection rate) while we report Recall@1---these coincide for single-tool queries but may differ for multi-tool ones; (2)~LLM-based selection uses full query text with in-context descriptions, while retrieval compresses tools into fixed embeddings; (3)~LLMs can handle novel tools zero-shot, whereas OATS needs outcome data to improve past the static baseline.

Other benchmarks tell a similar story: AppSelectBench~\citep{chen2025appselectbench} reports GPT-5 at 63.3\% on 100+ desktop tools, and GTA~\citep{wang2024gta} found GPT-4 completing fewer than 50\% of real-world tool-use tasks. LLM-based tool selection is both slow and unreliable; lightweight retrieval with outcome refinement offers a better cost--accuracy tradeoff for routing.

\begin{table}[h]
\centering
\caption{Selection performance on ToolBench (2,413 tools, 180 test queries) and MetaTool (199 tools, 1,287 test queries). Same 30\% test split for all methods. Best in \textbf{bold}, second \underline{underlined}.}
\label{tab:main}
\begin{tabular}{l cccc cccc}
\toprule
& \multicolumn{4}{c}{\textbf{ToolBench}} & \multicolumn{4}{c}{\textbf{MetaTool}} \\
\cmidrule(lr){2-5} \cmidrule(lr){6-9}
\textbf{Method} & R@1 & R@3 & R@5 & NDCG@5 & R@1 & R@3 & R@5 & NDCG@5 \\
\midrule
Random       & 0.238 & 0.656 & 0.829 & 0.692 & 0.096 & 0.308 & 0.492 & 0.298 \\
BM25         & \underline{0.392} & 0.772 & \underline{0.921} & \underline{0.853} & 0.397 & 0.617 & 0.753 & 0.595 \\
SE           & 0.382 & 0.765 & 0.897 & 0.834 & 0.716 & 0.904 & 0.958 & 0.869 \\
SE + Lexical & \textbf{0.388} & \underline{0.783} & \textbf{0.919} & \textbf{0.854} & 0.640 & 0.860 & 0.927 & 0.816 \\
\midrule
OATS-S1      & 0.381 & \textbf{0.793} & 0.915 & 0.848 & \textbf{0.830} & \textbf{0.964} & 0.986 & \textbf{0.940} \\
OATS-S2      & 0.372 & 0.774 & 0.896 & 0.823 & 0.716 & 0.909 & 0.954 & 0.869 \\
OATS-S3      & 0.387 & 0.778 & 0.906 & 0.841 & \underline{0.810} & \underline{0.956} & \textbf{0.988} & \underline{0.931} \\
\bottomrule
\end{tabular}
\end{table}

\begin{table}[h]
\centering
\caption{Ablation: incremental cost and contribution of each OATS component. Same 30\% test split. Embedding refinement adds zero cost; the MLP re-ranker degrades ToolBench due to data sparsity ($<$0.15 examples/tool).}
\label{tab:ablation}
\begin{tabular}{l r r cc cc}
\toprule
& \textbf{Serving} & \textbf{Added} & \multicolumn{2}{c}{\textbf{ToolBench}} & \multicolumn{2}{c}{\textbf{MetaTool}} \\
& \textbf{params} & \textbf{latency} & NDCG@5 & $\Delta$ & NDCG@5 & $\Delta$ \\
\midrule
SE (baseline)                 & 0     & ---       & 0.834 & --     & 0.869 & --     \\
+ Emb.\ refinement (core)    & 0     & 0\,ms     & 0.848 & +0.014 & \textbf{0.940} & +0.071 \\
+ MLP re-ranker (ablation A)  & 2.6K  & $<$0.5\,ms & 0.823 & \textcolor{red}{-0.011} & 0.869 & +0.000 \\
+ Contrastive (ablation B)    & 197K  & $<$0.1\,ms & 0.841 & +0.007 & 0.931 & +0.062 \\
\bottomrule
\end{tabular}
\end{table}

\begin{table}[h]
\centering
\caption{Inference latency per request (CPU-only). All methods stay within single-digit millisecond p50.}
\label{tab:latency}
\begin{tabular}{l cc cc l}
\toprule
& \multicolumn{2}{c}{\textbf{MetaTool}} & \multicolumn{2}{c}{\textbf{ToolBench}} & \\
\cmidrule(lr){2-3} \cmidrule(lr){4-5}
\textbf{Method} & p50 (ms) & p99 (ms) & p50 (ms) & p99 (ms) & \textbf{GPU} \\
\midrule
SE (baseline)       & 3.66 & 6.47 & 4.59 & 9.83 & No \\
OATS-S1             & 3.53 & 5.99 & 4.42 & 6.19 & No \\
OATS-S2             & 3.92 & 6.91 & 5.04 & 16.38 & No \\
OATS-S3             & 3.87 & 7.25 & 4.95 & 11.11 & No \\
\bottomrule
\end{tabular}
\end{table}

\begin{figure}[h]
\centering
\includegraphics[width=0.48\textwidth]{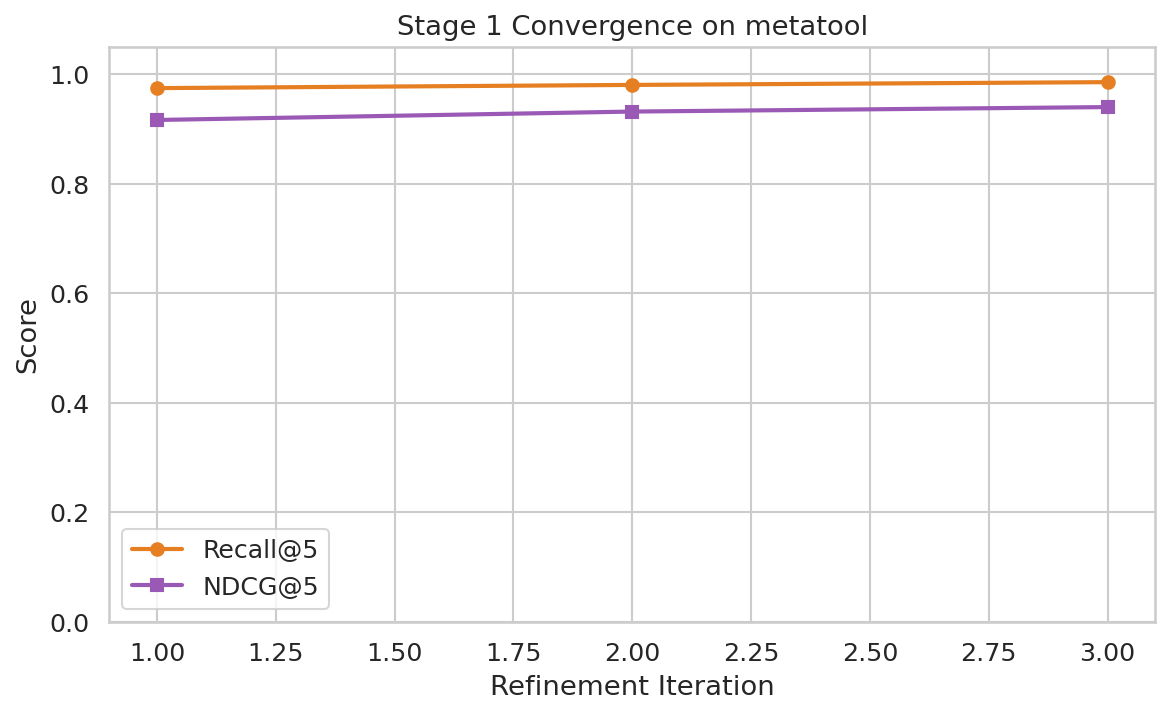}
\includegraphics[width=0.48\textwidth]{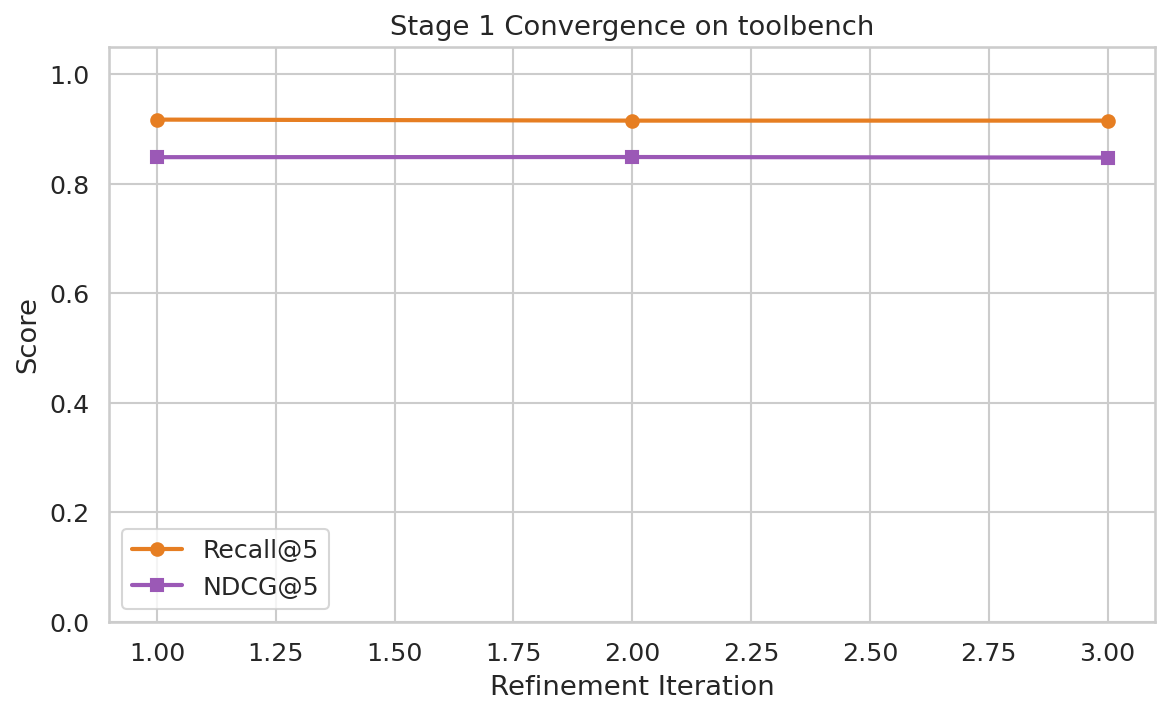}
\caption{Stage~1 convergence over iterations. Left: MetaTool. Right: ToolBench.}
\label{fig:convergence}
\end{figure}

\begin{figure}[h]
\centering
\includegraphics[width=0.48\textwidth]{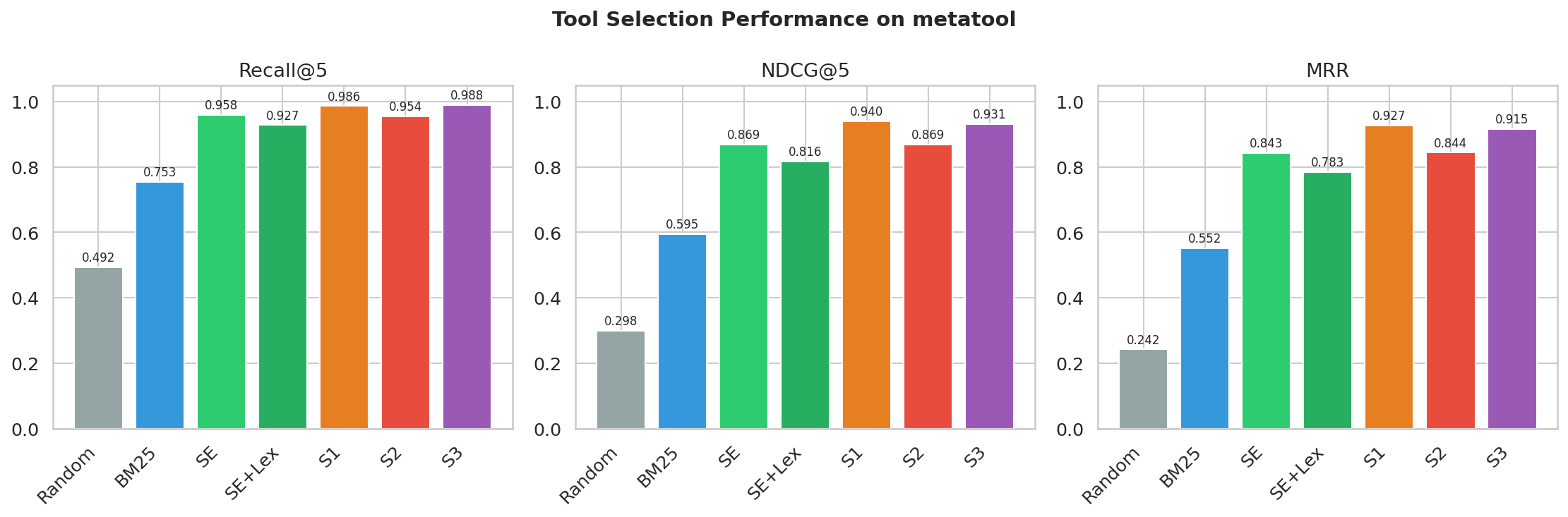}
\includegraphics[width=0.48\textwidth]{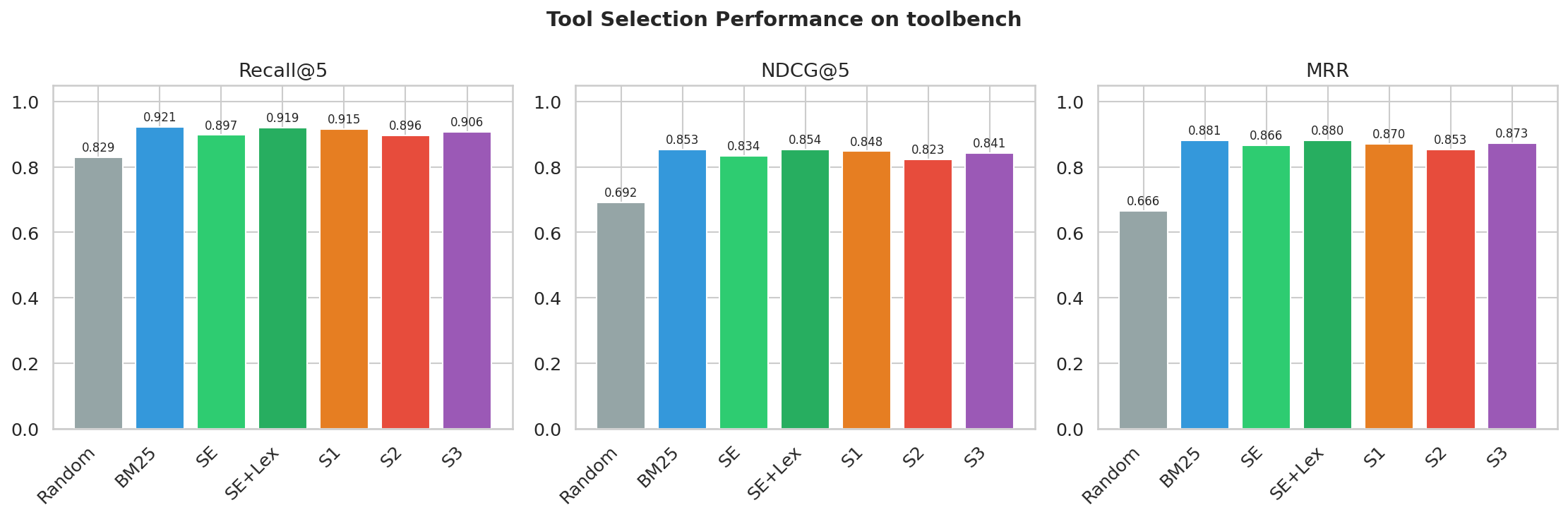}
\caption{Selection performance across all methods. Left: MetaTool. Right: ToolBench.}
\label{fig:results}
\end{figure}

\section{Discussion}
\label{sec:discussion}

\subsection{Why Not Use an LLM for Selection?}

One might ask whether the LLM itself could select its own tools. There are three problems with this:

\begin{enumerate}[leftmargin=*]
\item \textbf{Circular dependency.} Tools must be attached \emph{before} inference begins, so the LLM would need a separate pre-flight call, doubling latency and cost.

\item \textbf{Resource contention.} GPU capacity is the bottleneck in LLM serving. At 10K rps, an 8B selection model would consume 5--10 GPUs just for routing.

\item \textbf{Latency.} Tool selection happens before the request reaches any model pool. Adding 500\,ms of LLM inference on top of 200--2,000\,ms model inference increases total latency by 25--250\%, directly degrading time-to-first-token.
\end{enumerate}

OATS avoids all three by learning offline and deploying only a refined embedding table on CPU.

\subsection{Deployment Architecture}

OATS integrates into existing router infrastructure with minimal changes:

\begin{itemize}[leftmargin=*]
\item \textbf{Embedding refinement} runs as a cron job: read outcome logs, compute centroid updates, validate, and swap the embedding table. No code changes to the serving path.

\item \textbf{MLP re-ranker} (optional) adds a re-scoring call after \texttt{FindSimilarTools}. Gate behind a data-density check ($\geq 10$ examples/tool).

\item \textbf{Contrastive adapter} (optional) replaces the embedding model in \texttt{ToolsDatabase}. Output dimension is unchanged, so no downstream changes. Rollback is instantaneous.
\end{itemize}

\subsection{When to Add Complexity Beyond Refinement}

Based on our results:

\begin{itemize}[leftmargin=*]
\item $|\mathcal{T}| < 200$: Embedding refinement alone. Our MetaTool results (199 tools, +8.3\% NDCG) confirm this.

\item $|\mathcal{T}|$ = 50--500, $>$5K logs: Refinement + MLP re-ranker, but only if the data-to-tool ratio exceeds $\sim$10:1. Below that, it hurts.

\item $|\mathcal{T}| > 500$, $>$10K logs: Refinement + contrastive adapter. Contrastive learning scales better than pointwise re-ranking for large tool sets. Skip the MLP unless data is abundant ($> 50$ examples/tool).
\end{itemize}

\subsection{Limitations}

\paragraph{Outcome signal quality.} Our experiments use a binary proxy (match/no-match with ground truth). Production outcomes are richer---a tool might be selected but unused, used but erroring, or used but suboptimal. The framework accepts any scalar signal; richer signals should yield stronger improvements.

\paragraph{Cold start.} New tools have no outcome history and fall back to static similarity. They warm up as data accumulates.

\paragraph{Embedding model ceiling.} All stages are bounded by the base model's capacity. If \texttt{all-MiniLM-L6-v2} cannot distinguish two tools' semantic domains, refinement will not help. A larger base model would provide orthogonal gains.

\paragraph{Evaluation scope.} We measure selection quality, not end-to-end task completion. Good selection does not guarantee the LLM uses tools correctly, but it is a prerequisite: the right tool must be in the context window.

\section{Conclusion}
\label{sec:conclusion}

We presented OATS, which incorporates outcome feedback into tool selection by interpolating tool embeddings offline toward the centroid of queries where they succeed. This adds no parameters, latency, or GPU cost at serving time, yet improves NDCG@5 from 0.869 to 0.940 on MetaTool and from 0.834 to 0.848 on ToolBench, within single-digit millisecond CPU budgets.

The ablation study shows that learned components do not always help. The MLP re-ranker degrades accuracy when outcome data is sparse (below a $\sim$10:1 data-to-tool ratio), while the contrastive adapter provides modest but consistent gains for larger tool sets.

For production LLM routing, tool selection should be treated as a learned retrieval problem, but the first step should be the simplest one. A zero-cost embedding refinement, informed by production outcomes, captures most of the available gains. More complex components or LLM-based selection rarely justify their cost at the routing layer.

\bibliographystyle{plainnat}

\newpage
\appendix

\section{OATS-S1: Step-by-Step Walkthrough}
\label{app:examples}

We trace Algorithm~\ref{alg:s1} on a real MetaTool test query. All numbers are from the 70/30 train/test split with \texttt{all-MiniLM-L6-v2} embeddings.

\subsection{Setup: The Query and Candidate Tools}

\begin{quote}
\textbf{Test query:} ``Could you please search for and provide the complete and verbatim transcript of the strategy call that took place last week between ourselves and the executives?''
\end{quote}

\noindent The MetaTool benchmark provides 11 candidate tools for this query. The ground-truth answer is \texttt{buildbetter}, a meeting-transcript tool. Table~\ref{tab:walkthrough_cands} lists all candidates with their descriptions.

\begin{table}[h]
\centering
\caption{Candidate tools for the example query, with original descriptions from MetaTool.}
\label{tab:walkthrough_cands}
\small
\begin{tabular}{l p{8.5cm}}
\toprule
\textbf{Tool} & \textbf{Description} \\
\midrule
\texttt{buildbetter}$^\star$ & Chat with the knowledge of all your calls in BuildBetter (Zoom, GMeet, Webex). Start for free @ BuildBetter.ai \\
\texttt{QuiverQuantitative} & Access data on congressional stock trading, lobbying, insider trading, and proposed legislation. \\
\texttt{MixerBox\_WebSearch} & Search and summarize the web with our customized search engine powered by Google Search API! \\
\texttt{brandfetch} & Retrieve company and brand data including logos, colors, fonts, and other brand information. \\
\texttt{C3\_Glide} & Get live aviation data for pilots. Ask questions about METARs, TAFs, NOTAMs\ldots \\
\texttt{locator} & Displaying the current coordinates of the ISS and the names of the current crew. \\
\texttt{speechki\_tts} & The easiest way to convert texts to ready-to-use audio. \\
\texttt{MemoryTool} & A learning application with spaced repetition functionality. \\
\texttt{CTCP} & Analyze eligibility criteria in ClinicalTrials.gov. \\
\texttt{StrologyTool} & Provides astrology services for you. \\
\texttt{ExchangeTool} & Seamlessly convert currencies with our integrated currency conversion tool. \\
\bottomrule
\multicolumn{2}{l}{\footnotesize $^\star$\,Ground-truth tool.}
\end{tabular}
\end{table}

\noindent The correct tool's description says ``BuildBetter.ai''---a branded name that tells an embedding model nothing about meeting transcripts. The query mentions ``executives'' and ``strategy call,'' which have lexical affinity to business/finance tools.

\subsection{Step 1: Static Retrieval (Before Refinement)}

We encode the query and all 11 candidate descriptions with the base embedding model, then rank by cosine similarity:

\begin{center}
\small
\begin{tabular}{cl r}
\toprule
\textbf{Rank} & \textbf{Tool} & \textbf{Cosine sim} \\
\midrule
1 & \texttt{QuiverQuantitative} & 0.337 \\
\rowcolor{black!8}
2 & \texttt{buildbetter}$^\star$ & 0.276 \\
3 & \texttt{MixerBox\_WebSearch} & 0.254 \\
4 & \texttt{brandfetch} & 0.228 \\
5 & \texttt{C3\_Glide} & 0.200 \\
\multicolumn{3}{c}{\footnotesize\textit{(6 more candidates below 0.18)}} \\
\bottomrule
\end{tabular}
\end{center}

\noindent Static embedding puts the correct tool at rank~2. \texttt{QuiverQuantitative} wins because ``congressional\ldots lobbying\ldots legislation'' sits in a similar embedding region as ``strategy call\ldots executives.'' The margin is $0.337 - 0.276 = 0.061$.

\subsection{Step 2: Collecting Outcome Data (Algorithm~\ref{alg:s1}, Steps 1--2)}

From the 70\% training split, we collect the queries where \texttt{buildbetter} appears in the ground-truth relevant tools ($Q^+$) and queries where it was retrieved in the top-$K$ but was \emph{not} relevant ($Q^-$):

\paragraph{Positive queries $Q^+$ (4 unique training queries):}
\begin{enumerate}[leftmargin=2em, label=\textcolor{teal}{$+$\arabic*}., itemsep=1pt]
\item ``Retrieve the complete and unedited transcript of the customer support call I had with John Doe\ldots''
\item ``I am unable to locate the complete and accurate transcript of yesterday's standup meeting\ldots''
\item ``Can you find the complete and accurate transcript of the last sales call I had, specifically with Acme Corp\ldots''
\item ``I had a strategy meeting last week, what were the key points we discussed?''
\end{enumerate}

\paragraph{Hard-negative queries $Q^-$ (7 training queries where \texttt{buildbetter} was retrieved but wrong):}
\begin{enumerate}[leftmargin=2em, label=\textcolor{red!70!black}{$-$\arabic*}., itemsep=1pt]
\item ``Can the CLINQ plugin display the detailed information of the total duration for all your calls\ldots''
\item ``Can you help me add a new page to my Notion workspace?''
\item ``Can you create a recurring call reminder for my weekly team meetings on CLINQ?''
\end{enumerate}

\noindent The positive queries cluster around ``retrieving transcripts from meetings and calls.'' The negatives mention ``calls'' but are about call management (CLINQ) or unrelated tools---cases where \texttt{buildbetter} was a false positive.

\subsection{Step 3: Computing the Refined Embedding (Algorithm~\ref{alg:s1}, Step 3)}

We encode all $Q^+$ queries, compute their centroid $\bar{e}^+$, and do the same for $Q^-$. Then we apply the update rule from Equation~\ref{eq:desc_refine}:

\begin{equation*}
\hat{e}_{\texttt{buildbetter}} = \underbrace{0.7 \cdot e(\texttt{buildbetter})}_{\text{retain original}} \;+\; \underbrace{0.3 \cdot \bar{e}(Q^+)}_{\text{attract toward transcripts}} \;-\; \underbrace{0.1 \cdot \bar{e}(Q^-)}_{\text{repel from call-management}}
\end{equation*}

\noindent\textbf{Geometric interpretation.} The original embedding sits in a generic ``productivity SaaS'' region. The positive centroid $\bar{e}(Q^+)$ sits in a ``meeting transcripts'' region. The interpolation pulls the tool toward transcripts and pushes it away from call management. After re-normalization, the tool lives where transcript-retrieval queries concentrate.

\subsection{Step 4: Re-Ranking After Refinement}

With the refined \texttt{buildbetter} embedding (and refined embeddings for all other tools that had training data), we re-rank the same 11 candidates:

\begin{center}
\small
\begin{tabular}{cl rr}
\toprule
\textbf{Rank} & \textbf{Tool} & \textbf{New sim} & \textbf{$\Delta$} \\
\midrule
\rowcolor{teal!10}
1 & \texttt{buildbetter}$^\star$ & 0.440 & \textbf{+0.164} \\
2 & \texttt{QuiverQuantitative} & 0.343 & +0.005 \\
3 & \texttt{MixerBox\_WebSearch} & 0.283 & +0.029 \\
4 & \texttt{brandfetch} & 0.253 & +0.026 \\
5 & \texttt{C3\_Glide} & 0.243 & +0.043 \\
\multicolumn{4}{c}{\footnotesize\textit{(6 more candidates below 0.16)}} \\
\bottomrule
\end{tabular}
\end{center}

\noindent The correct tool moves from rank~2 to rank~1, gaining $+0.164$ in similarity. \texttt{QuiverQuantitative} barely moves ($+0.005$) because its training signal (congressional trading queries) is unrelated to transcripts. The margin reverses from $-0.061$ to $+0.097$.

\paragraph{Why the correction is large.} \texttt{buildbetter} had only 4~positive and 7~negative training queries. But the positives cluster tightly around ``call transcript retrieval,'' giving a directional centroid. The original embedding was far from this cluster because the description (``BuildBetter.ai'') is opaque. This is where refinement helps most: when the description fails to communicate function, outcome data fills the gap.

\subsection{Additional Examples}
\label{app:more_examples}

Table~\ref{tab:more_examples} summarizes three additional cases where OATS-S1 corrects a static embedding failure, spanning different MetaTool subtask types.

\begin{table}[h]
\centering
\caption{Additional OATS-S1 corrections on MetaTool test queries. ``SE sim'' and ``S1 sim'' are cosine similarities to the \emph{correct} tool before and after refinement. ``SE top-1'' is the wrong tool selected by static embedding.}
\label{tab:more_examples}
\small
\begin{tabular}{p{4.2cm} l l r r}
\toprule
\textbf{Query (truncated)} & \textbf{Correct} & \textbf{SE top-1} & \textbf{SE} & \textbf{S1} \\
\midrule
``Show me diverse and amusing memes\ldots'' & \texttt{lsongai} & \texttt{PolishTool} & 0.287 & \textbf{0.450} \\
``Discount codes for booking hotels or flights?'' & \texttt{Discount} & \texttt{TripTool} & 0.585 & \textbf{0.691} \\
``Meaning of \emph{arigatou} in Japanese?'' & \texttt{speak} & \texttt{ResearchFinder} & 0.077 & \textbf{0.149} \\
\bottomrule
\end{tabular}
\end{table}

\paragraph{Memes query.} \texttt{PolishTool} (``creative inspiration'') outscores \texttt{lsongai} (``AI-powered content\ldots memes'') by $+0.066$ under SE. After refinement, \texttt{lsongai} gains $+0.163$ from entertainment-related training queries. Failure mode: semantic decoy---similar descriptions, different function.

\paragraph{Discount codes query.} \texttt{TripTool} mentions ``discounted hotel,'' creating lexical overlap ($+0.057$ margin). Outcome data pulls \texttt{Discount} toward coupon/promo queries, flipping the margin to $+0.015$. Failure mode: lexical overlap masking functional intent.

\paragraph{Japanese translation query.} All candidates score below $0.15$---near random. ``speak'' does not embed well for translation queries. Training queries about language learning nearly double its similarity ($+0.072$). Failure mode: opaque name in a low-similarity regime.

\subsection{Summary of Failure Patterns}

Three recurring patterns emerge:
\begin{enumerate}[leftmargin=*]
\item \textbf{Semantic decoys}: Similar descriptions, different function. The positive centroid separates them by pulling each tool toward its actual use cases.
\item \textbf{Opaque descriptions}: Generic or branded names (``BuildBetter.ai,'' ``Speak'') that embeddings cannot interpret. Outcome data compensates.
\item \textbf{Low-similarity regimes}: When all candidates score similarly, small centroid interpolations flip rankings.
\end{enumerate}

\end{document}